\documentclass[letterpaper]{article} 
\usepackage{aaai2027}
\usepackage[hyphens]{url}  
\usepackage{graphicx} 
\usepackage{natbib}  
\usepackage{caption} 
\usepackage{algorithm}
\usepackage{algorithmic}
\usepackage{amsmath}
\usepackage{amssymb}
\usepackage{multirow}
\usepackage{booktabs}
\usepackage[table]{xcolor}

\colorlet{bestred}{red!60}
\colorlet{secondred}{red!30}
\colorlet{thirdred}{red!10}

\newcommand{\best}[1]{\cellcolor{bestred}\textbf{#1}}
\newcommand{\second}[1]{\cellcolor{secondred}#1}
\newcommand{\third}[1]{\cellcolor{thirdred}#1}
\usepackage{newfloat}
\usepackage{listings}
\DeclareCaptionStyle{ruled}{labelfont=normalfont,labelsep=colon,strut=off} 
\floatstyle{ruled}
\newfloat{listing}{tb}{lst}{}
\floatname{listing}{Listing}

\usepackage{booktabs}

\title{Diagnosing and Mitigating Perception-Decision Misalignment in Omni-LLMs via Modality Subspace Activation}

\author{
    Hongbo Jiang\textsuperscript{\rm 1},
    Jie Li\textsuperscript{\rm 2}\equalcontrib,
    Yunhang Shen\textsuperscript{\rm 3},
    Tianyu Xie\textsuperscript{\rm 1},
    Pingyang Dai\textsuperscript{\rm 1}\corresponding
}

\affiliations{
    \textsuperscript{\rm 1}Xiamen University\\
    \textsuperscript{\rm 2}Shanghai Artificial Intelligence Laboratory\\
    \textsuperscript{\rm 3}Tencent YouTu Laboratory
}

\begin{document}

\maketitle

\begin{abstract}
Omni-Large Language Models (Omni-LLMs) have emerged as the foundation for complex multi-modal reasoning, powering many research and application areas such as World Action Models (WAMs) and autonomous agents. 
However, their leading task performance often masks a critical issue: a profound \textbf{Perceptual-Decision Misalignment (PDM)}, where strategic decisions remain unfaithful to multi-modal perceptions. 
To systematically assess this structural gap, we formalize \textbf{Causal Modality Sensitivity (CMS)} as a diagnostic perspective to gauge the degree of perceptual-decision misalignment in omni-modal inference.
We operationalize CMS through a dual-lens metric framework: \textbf{Answer Retention Rate (ARR)} at the macro behavioral level, and \textbf{Logit Angular Discrepancy (LAD)}, a continuous geometric metric that mathematically maps microscopic distribution shifts within the multi-choice space. 
Furthermore, we curate \textbf{CausalMSBench}, a diagnostic dataset that strictly isolates language priors and enforces multi-modal dependency.
Our benchmarking reveals that the popular Omni-LLMs exhibit a critically low CMS, showing almost negligible distribution shifts within the multi-choice space even when indispensable modalities are entirely omitted. 
This phenomenon demonstrates a systemic perceptual-decision misalignment rather than effective multi-modal integration. 
To rectify this, we propose \textbf{Modality Subspace Activation (MSA)}, a training-free inference-time framework that utilizes Singular Value Decomposition (SVD) to estimate modal activation strengths. 
MSA dynamically suppresses the projection of the stronger modality onto the last hidden state and boosts that of the weaker modality.
Extensive evaluations show that MSA consistently optimizes both ARR and LAD, effectively restoring the model's CMS. 
Ultimately, our work establishes a unified framework spanning from CMS diagnosis to inference-time mitigation for perceptual-decision misalignment in Omni-LLMs.
\end{abstract}



\section{Introduction}

The rapid advancement of Omni-Large Language Models (Omni-LLMs)\cite{xu2025qwen3omnitechnicalreport, xu2025qwen25omnitechnicalreport, ye2025omnivincienhancingarchitecturedata, tong2025interactiveomniunifiedomnimodalmodel}, empowered by their cross-modal processing capabilities, has established them as the foundation across critical artificial intelligence frontiers. 
This paradigm shift is particularly pronounced in areas such as World Action Models(WAMs) and autonomous agents\cite{cen2025worldvlaautoregressiveactionworld, li2024visionlanguagefoundationmodelseffective, brohan2023rt1roboticstransformerrealworld, wen2025diffusionvlageneralizableinterpretablerobot, black2026pi0visionlanguageactionflowmodel, huang2024embodiedgeneralistagent3d, xu2026agenticactiveomnimodalperception, zhu2026omniragagentagenticomnimodalreasoning}, where Omni-LLMs increasingly serve as the core behavioral and reasoning backbone.
However, these system frameworks still suffer from a critical issue: their strategic decisions are not necessarily faithful to their multi-modal perceptions.\cite{yuan2026quackquestioningunderstandingauditing, andrade2026letsthinkstepsmitigating}  
This issue is clearly evidenced by a key empirical observation: completely omitting indispensable sensory modalities may yield minimal variation in the model's final decision outputs.

\begin{figure}[t]
    \centering
    \includegraphics[width=\columnwidth]{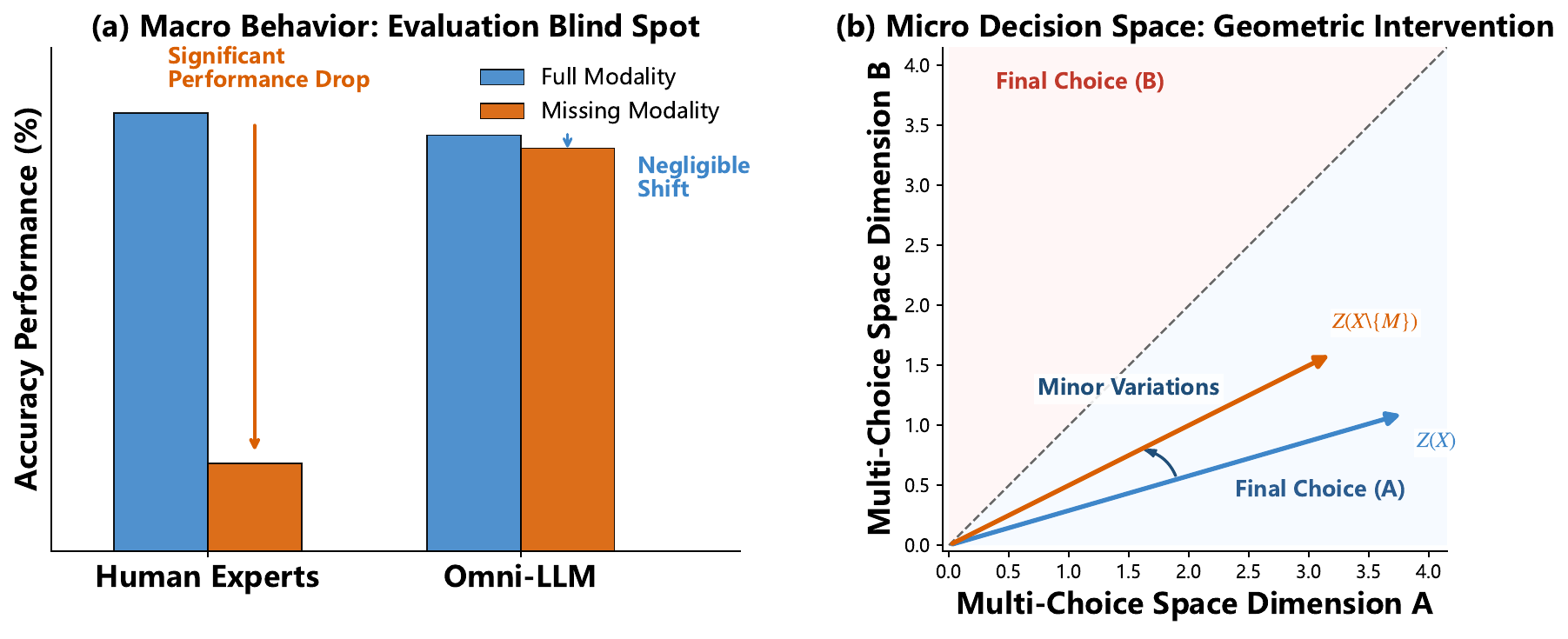}
    \caption{\textbf{Illustration of modality shortcut bottleneck and our geometric intervention framework.} 
(a) \textbf{Macro Behavior}: Unlike humans who heavily rely on modality $M$ (sharp drop upon omission), Omni-LLMs exhibit negligible performance shifts. 
(b) \textbf{Micro Decision Space}: When necessary modalities are omitted, the representation distribution undergoes only a minor variation in the multi-choice space.}
    \label{fig:motivation}
\end{figure}


As illustrated in the left panel of Figure~\ref{fig:motivation}, when indispensable input modalities are removed, the model’s accuracy and the outputs remain strikingly unperturbed. 
To investigate the potential causes of this misalignment, we carefully deconstruct the phenomenon as follows.
First, we consider the hypothesis of ``language priors''~\cite{ tan2025prospectretrospectreflectivememory, pan2025memoryconstructionretrievalpersonalized}.
When a textual query triggers strong language priors or parametric memories (e.g., when tapping into pre-trained factual knowledge~\cite{wu2025generalcontinuousmemoryvisionlanguage}), the model can directly retrieve the answer, bypassing perceptual reasoning and rendering non-text modalities redundant.
Second, another pivotal confounding factor is multi-modal dependency~\cite{chao2026jointavbench}, whether a targeted modality is strictly indispensable for answering the specific question by design. 
For instance, in a query like ``What is the voice of the woman in the audio shouting?'', the task design intrinsically dictates the redundancy of the visual stream and ``the question can be answered using vision or audio alone''~\cite{chao2026jointavbench}.
Crucially, when an evaluation dataset minimizes language priors and strictly enforces multi-modal dependency, any persistent output invariance under modality perturbations strongly signals, a profound \textbf{Perceptual-Decision Misalignment (PDM)}, where the final decisions are rendered largely in isolation from effective perceptual grounding.


While prior studies have identified issues like language priors or modality biases\cite{yan2026beyond, chaubey2026mod, kwon2025seesawmodalitybalancegradient, chen2024we}, their evaluations predominantly rely on macro-level metrics (e.g., accuracy)\cite{yan2026beyond, wu2025language, chao2026jointavbench, chaubey2026mod, gat2021perceptualscoredatamodalities}, which limits their ability to capture fine-grained behaviors.
After a modality perturbation, the model may still stick to its original answer, but its distribution within the multi-choice space can have changed dramatically.
To address this observational limitation, we formalize \textbf{Causal Modality Sensitivity (CMS)}, defined as the systematic output divergence of a model in response to the perturbations in input modalities.
To operationalize CMS, we engineer a dual-lens metric framework composed of \textbf{Answer Retention Rate (ARR)} and \textbf{Logit Angular Discrepancy (LAD)}. 
Under modality perturbations, ARR measures the model's persistence in retaining its original prediction, where a higher score indicates that modality perturbations have less impact on the final answer. 
LAD captures the distribution shifts within the multi-choice space, where a smaller value reflects a smaller impact on the model's option-level decision-making. 
Recognizing the equal importance of both macro-level and micro-level dimensions, the final CMS metric is formulated as their aggregated average.
To substantiate this evaluation, we construct \textbf{CausalMSBench}, a benchmark dataset curated to minimize language priors while strictly enforcing multi-modal dependency.
Using CausalMSBench and CMS, we observe a consistent phenomenon across existing models: while popular models achieve moderate task performance, they exhibit surprisingly low CMS scores. 
Specifically, under modality perturbations, many cases show that many models maintain a high ARR alongside substantial variations in LAD. 
This discrepancy indicates that decisions are made largely without effective perceptual grounding.


To mitigate the issue, conventional paradigms primarily resort to logit-level contrastive decoding\cite{leng2024mitigating, chen2025decouplingcontrastivedecodingrobust, chen2026maskmattersmitigatingobject}, attention-guided decoding\cite{guan2026thinkomni, xu-etal-2025-mitigating, liao2025rewardguidedspeculativedecodingefficient}, or prompt-based chains of thought that mandate detailed self-generated descriptions prior to the final answer\cite{feng2025lookreciteanswerenhancing, ghosh2025visualdescriptiongroundingreduces}. 
We propose \textbf{Modality Subspace Activation (MSA)}, a training-free, test-time hidden-state intervention framework. 
By collecting the model's last hidden state across controlled modality inputs, MSA constructs activation matrix of each modality and applies Singular Value Decomposition (SVD) to isolate the directional components and construct the corresponding subspace.
During real-time inference, MSA maps the inference-time hidden states onto these isolated subspaces, computing the $L_2$ norms of the projections to quantify the activation strength of each modality stream and sorting the modalities by activation strength. 
Upon detecting representation imbalances (e.g., when dominant language priors overshadow essential multi-modal cues), MSA adaptively executes a dual-direction geometric regulation. 
It suppresses the over-activated and stronger modality while boosting the weaker modality, thereby dynamically balancing the feature distribution across modalities and preventing dominant modalities from overriding the decision process.
With MSA, the evaluated models effectively reduced ARR and improved LAD, indicating that MSA successfully enhances the models' CMS.

In summary, our primary contributions are threefold:

\begin{itemize}
    \item \textbf{A Diagnostic Perspective}: We approach the assessment of Omni-LLMs through the lens of Perceptual-Decision Misalignment. 
    Under this perspective, we formalize \textbf{Causal Modality Sensitivity (CMS)} as a diagnostic evaluation dimension to quantify the alignment gap between perceptual multi-modal inputs and final decisions.
    \item \textbf{A Metric and Discoveries}: We propose \textbf{Logit Angular Discrepancy (LAD)}, a continuous geometric metric capable of detecting fine-grained distribution shifts within the multi-choice space. 
    Utilizing LAD alongside \textbf{Answer Retention Rate (ARR)}, we uncover a widespread phenomenon of perceptual-decision misalignment across a broad spectrum of mainstream Omni-LLMs.
    \item \textbf{A Training-Free Intervention Method}: We develop \textbf{Modality Subspace Activation (MSA)}, a plug-and-play, inference-time intervention mechanism. 
    By executing hidden state enhancement through modal subspace analysis, MSA effectively couples final decisions with perceptual features without parameter retraining.
\end{itemize}


\section{Related Work}

\subsection{Perceptual-Decision Misalignment}
The alignment between cross-modal perception and strategic reasoning has emerged as a crucial benchmark for the reliability of Omni-Large Language Models (Omni-LLMs).
QUACK\cite{yuan2026quackquestioningunderstandingauditing} proposes an auditing and evaluation framework for the ``perception-decision'' consistency of multi-modal agents in long-term dialogs, revealing the ``perception-expression disconnect'' of frontier VLMs in long-range memory and dynamic confrontation. 
To tackle the hallucination in which models arrive at correct answers based on faulty visual understanding, PaLMR\cite{li2026palmrfaithfulvisualreasoning} introduces a process-aligned framework.
Furthermore, SGV\cite{andrade2026letsthinkstepsmitigating} presents a ``self-insulated verification'' method to mitigate ``desirability bias'', the strong tendency of MLLMs to misjudge digital agent trajectories. 
These approaches evaluate the relationship between perception and decision, jointly highlighting the criticality of ``perception-decision'' alignment.

\subsection{Modality Hallucinations in Omni-LLMs}

Previous literature predominantly attributes this phenomenon to language priors or modality preferences. 
Specifically, Wu et al.~\cite{wu2025language} focuses on textual priors, systematically analyzing the attention allocation of various MLLMs across different modalities and revealing that the average attention weight dedicated to text tokens severely dominates over other modalities.
Furthermore, MMStar~\cite{chen2024we} directly removes the visual modality entirely, only to find a marginal degradation in the models' retrieval accuracy. 
Expanding beyond text dominance, Yan et al.~\cite{yan2026beyond} establishes a tri-modal semantic conflict evaluation framework to quantify which modality the model trusts more implicitly, thereby gauging the degree of modality preference.
However, these studies\cite{chen2024we, yan2026beyond, gat2021perceptualscoredatamodalities}
largely rely on macro-level metrics such as accuracy to quantify hallucinations in Omni-LLMs. 
To remedy the deficiency in fine-grained analysis, we introduce \textbf{Logit Angular Discrepancy (LAD)} as a microscopic metric, alongside \textbf{Answer Retention Rate (ARR)} to complement the macroscopic diagnosis dimension. 
In terms of dataset curation, to minimize language priors while strictly enforcing multi-modal dependency, our workflow heavily builds upon the footprints of JointAVBench~\cite{chao2026jointavbench}. 
Each sample within this evaluation dataset exhibits a strict multi-modal dependency where all three modalities are indispensable during inference.
Hence, we integrate it as a core component of our newly proposed \textbf{CausalMSBench}.

\subsection{Training-Free Method to Mitigate Omni-Modal Hallucinations}

Existing training-free methodologies for mitigating Omni-Modal hallucinations can be systematically categorized into four paradigms. 
The first is contrastive decoding, pioneered by works like VCD\cite{leng2024mitigating}, which computes the output logits under standard inputs alongside the corrupted logits derived from distorted visual perturbations, deploying a contrastive formulation to adaptively calibrate the decoding weights between the two streams. 
The second is guided decoding, exemplified by ThinkOmni\cite{guan2026thinkomni}, which inherently operates on a contrastive mechanism but utilizes the execution trajectories of Large Reasoning Models (LRMs) as a reference to rectify reasoning-induced hallucinations within Omni-LLMs. 
The third involves description-based prompting paradigms, such as ``Look, Recite, Then Answer''\cite{feng2025lookreciteanswerenhancing}, which decomposes VLM inference into a three-stage pipeline of searching (generating visual descriptions), reciting (triggering internal LLM knowledge via a router), and answering (cross-verifying the alignment between descriptions and knowledge). 
The final paradigm focuses on hidden-state representation modification: 
``Steer Where It Matters''~\cite{zhang2026steermatterstokenlevelvisualsensitivity} reveals that the impact of visual conditioning on token prediction is sparse and heavy-tailed, thereby advocating for a fine-grained adaptive steering restricted solely to ``visually sensitive'' tokens.
CausalLens\cite{ji2026causallens} monitors the attention concentration of different heads on image tokens to identify ``visually reliable heads'', augmenting their visual contributions in intermediate layers while suppressing system prompts and textual priors.
Our proposed \textbf{MSA (Modality Subspace Activation)} leverages the intrinsic properties of CausalMSBench and builds upon the microscopic empirical insights unveiled by LAD. 
By introducing systematic modality perturbations, MSA quantifies the activation strength of each modality with full-modality inputs, ultimately executing a ``suppressing the stronger while boosting the weaker'' enhancement scheme directly on the last hidden state to prevent modality dominance and promote balanced representation utilization.


\begin{figure*}[t]
    \centering
    \includegraphics[width=\textwidth]{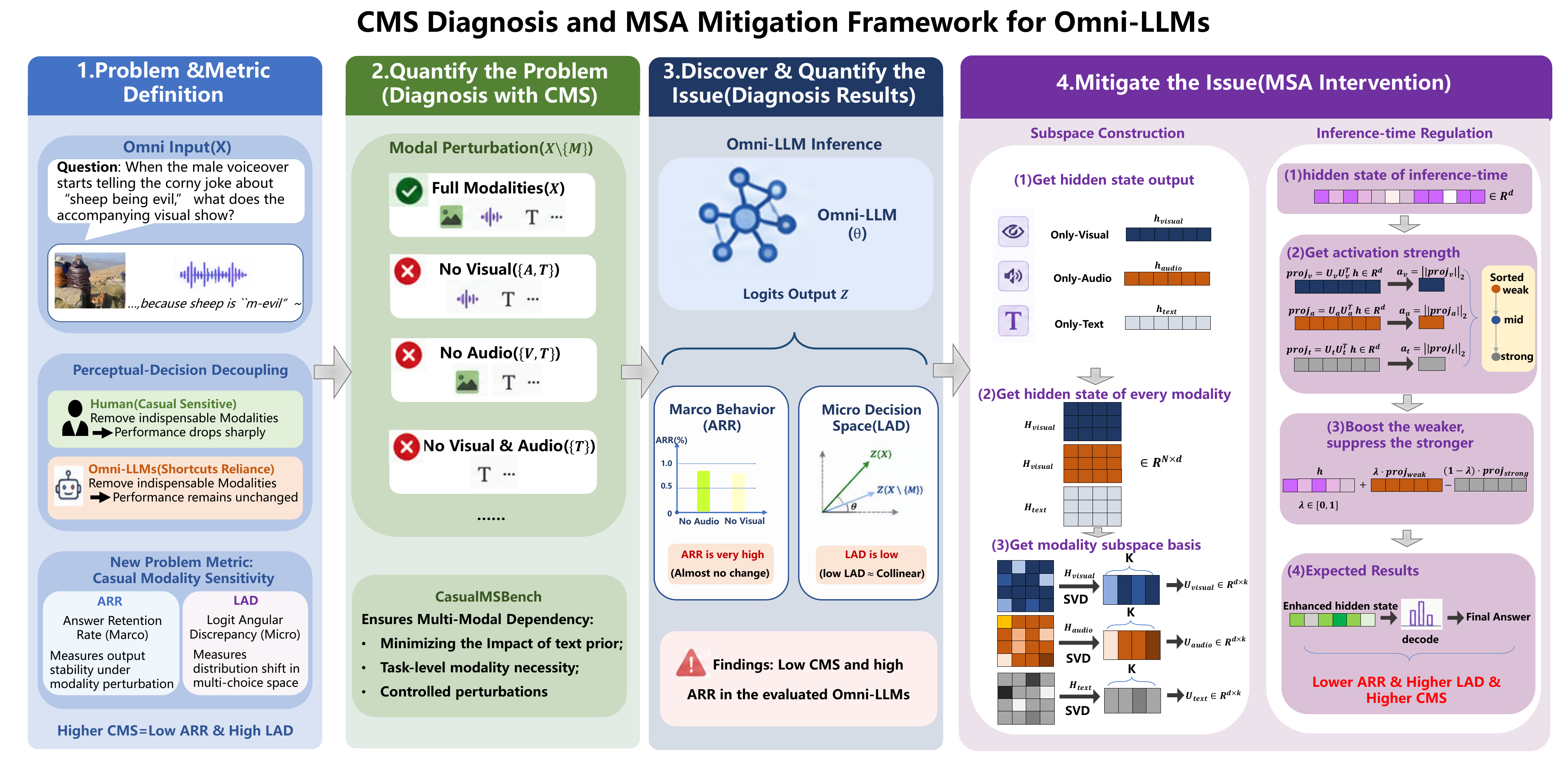}
    \caption{\textbf{Overview of our proposed CMS Diagnostic and MSA Mitigation Framework for Omni-LLMs.} 
    \textbf{(1) Benchmark \& Metric Formulation}: CausalMSBench enforces multi-modal dependency and introduces Causal Modality Sensitivity (CMS) via Macro Answer Retention Rate (ARR) and Micro Logit Angular Discrepancy (LAD). 
    \textbf{(2) Quantification \& Inference}: Evaluate models under controlled modality perturbations ($X \setminus \{M\}$) to diagnose potential Perception-Decision Misalignment (PDM). 
    \textbf{(3) Diagnostic Findings}: Omni-LLMs exhibit high ARR and low LAD under perturbations, confirming severe PDM insensitivity compared to human causal behavior. 
    \textbf{(4) Modality Subspace Activation (MSA)}: MSA mitigates PDM at inference time by constructing modal activation SVD bases ($U_M$) from representative samples, dynamically boosting weak modalities while suppressing dominant ones.}
    \label{fig:framework}
\end{figure*}

\section{Method}
\subsection{Multi-Modal Dependency}
Let $M = \{V, A, T\}$ denote the full set of available modalities, representing Visual, Audio, and Text modalities, respectively. 
We define $M' \subsetneq M$ as any proper subset of $M$, representing a scenario with missing or incomplete modalities. 
Consider an Omni-Modal evaluation dataset $D = \{(x_i, y_i^*)\}_{i=1}^N$, where $x_i$ is the multi-modal input sample and $y_i^*$ is the corresponding ground-truth answer. 
To quantify the behavioral correctness of a model, we define a matching indicator function $\mathbb{U}_{\text{match}}(y, y^*)$, which outputs 1 if the predicted answer $y$ perfectly matches the ground truth $y^*$ (Correct Event), and 0 otherwise (Failure Event).

To ensure that the dataset $D$ rigorously evaluates the perceptual-decision misalignment, we impose a strict constraint validated by a pool of human experts $H$. 
For any input of a sample $x_i \in D$ and any human evaluator $h_j \in H$, the gap between the perception accuracy under full modalities and the maximum perception accuracy under any modal-incomplete input must satisfy the following constraint:
\begin{equation}
    \begin{aligned}
        & P(\text{correct} \mid x_i, M, h_j) - \max_{M' \subsetneq M} P(\text{correct} \mid x_i, M', h_j) \\
        & \qquad \ge \delta, \quad \forall x_i \in D, \ \forall h_j \in H,
    \end{aligned}
\end{equation}
where $\delta > 0$ represents a predefined minimum acceptable perceptual gap threshold. 
When a dataset $D$ satisfies this condition, we define it as strictly `` \textbf{Multi-Modal Dependency}''. 
This implies that for any sample within this dataset, \textbf{the omission of any indispensable modality will inevitably lead to a severe degradation in human performance}.

\subsection{Causal Modality Sensitivity}
With a dataset exhibiting Multi-Modal Dependency established, we formalize the behavioral analysis of an Omni-Modal Large Language Model (Omni-LLM) $f$. 
When the model $f$ receives an input $x_i$, its conditional probability distribution over output candidates is denoted as $P_f(\cdot \mid x_i)$. 
The \textbf{Causal Modality Sensitivity (CMS)} of the model can be evaluated by the expected behavioral divergence under full-modality versus modal-omission conditions. 
If a model evaluated on a Multi-Modal Dependency dataset exhibits:
\begin{equation}
    \begin{split}
        \mathbb{E}_{D} \left[ \left\vert{} P_f(\text{correct} \mid x_i, M) - P_f(\text{correct} \mid x_i, M') \right\vert{} \right] \le \gamma,
    \end{split}
\end{equation}
where $\gamma > 0$ is an acceptable upper bound, it indicates that the model lacks basic awareness regarding the absence of essential modalities, thereby demonstrating a low CMS.

The macro-level metrics based on final token outputs or discrete accuracy drop suffer from an inherent limitation: they treat the model's decision-making process as a black box, failing to capture the microscopic distribution within the multi-choice space. 
To overcome this, we propose the geometric metric \textbf{Logit Angular Discrepancy (LAD)}.
Let $\mathbf{h}(X)$ and $\mathbf{h}(X \setminus \{M_i\})$ represent the top-layer hidden states obtained under the full-modality input $X$ and the modality-omitted input $X \setminus \{M_i\}$ ($M_i \in \{V, A\}$), respectively. Passing these representations through the language model head yields the corresponding raw logit vectors.
To eliminate background noise from irrelevant tokens and focus strictly on the decision boundary, we extract the logits restricted to the option space $\mathcal{O}_\text{options}$ (e.g., choices A, B, C, D), denoted as $\mathbf{z}(X)$ and $\mathbf{z}(X \setminus \{M_i\})$. 
We then zero-center these option logits to capture the pure shift in relative preference, defined as $\tilde{\mathbf{z}} = \mathbf{z} - \bar{\mathbf{z}}$, where $\bar{\mathbf{z}}$ is the mean scalar across the options.
The LAD is then defined as the angular displacement between these zero-centered option logit vectors:
\begin{equation}
\begin{aligned}
\text{LAD}(X, M_i)
=& 1 - \text{CosineSim}\bigl(\tilde{\mathbf{z}}(X), \tilde{\mathbf{z}}(X \setminus {M_i})\bigr) \\
=& 1 - \frac{\tilde{\mathbf{z}}(X) \cdot \tilde{\mathbf{z}}(X \setminus {M_i})}{\Vert \tilde{\mathbf{z}}(X)\Vert_2 ,\Vert \tilde{\mathbf{z}}(X \setminus {M_i})\Vert_2}.
\end{aligned}
\end{equation}

Through LAD, we can bypass the camouflage of macroscopic accuracy and track the subtle logit discrepancies within the model's internal representation space at a microscopic level.
To complement LAD with a discrete macro-behavioral counterpart, we further introduce the \textbf{Answer Retention Rate (ARR)}, which operationalizes the absolute rigidity of the model's final discrete choices under modality omission.
The hard top-1 predicted options under the full-modality and missing-modality settings are extracted via the argmax operator over the sliced logit vectors, respectively:
\begin{equation}
    \begin{aligned}
        &\hat{y}(X) = \arg\max_{o \in \mathcal{O}_\text{options}} \mathbf{z}_o(X),\\
        &\hat{y}(X \setminus \{M_i\}) = \arg\max_{o \in \mathcal{O}_\text{options}} \mathbf{z}_o(X \setminus \{M_i\}).
    \end{aligned}
\end{equation}
At the individual sample level, the answer retention status is modeled as a binary event $\text{AR}(X, M_i) \in \{0, 1\}$. 
It evaluates to $1$ (True) if the model's discrete choice remains invariant despite the omission of modality $M_i$, and $0$ (False) if the perturbation successfully forces a prediction flip:
\begin{equation}
    \text{AR}(X, M_i) = \mathbb{I}\left( \hat{y}(X) = \hat{y}(X \setminus \{M_i\}) \right),
\end{equation}
where $\mathbb{I}(\cdot)$ denotes the standard indicator function. 
Aggregating across the entire evaluation dataset $\mathcal{D}$, the macroscopic ARR is defined as the statistical expectation of this decision invariance:
\begin{equation}
    \text{ARR}(\mathcal{D}, M_i) = \frac{1}{|\mathcal{D}|} \sum_{X \in \mathcal{D}} \text{AR}(X, M_i).
\end{equation}
By pairing the discrete ARR with the continuous LAD, our framework establishes a multi-granularity diagnostic lens capable of simultaneously capturing both macroscopic decision flips and microscopic distribution shifts.

To synthesize these complementary perspectives into a unified scalar measure, we formally define CMS as the average of the macro-level choice alteration (i.e., $1 - \text{ARR}$) and the micro-level logit displacement:
\begin{equation}
\text{CMS} = \frac{(1 - \text{ARR}) + \text{LAD}}{2}.
\end{equation}
A lower score reflects that the model's outputs are insensitive to perturbations of indispensable modalities, which further indicates the misalignment between perception and decision.

\subsection{Modality Subspace Activation}
To rectify the issue of misalignment between perception and decision-making, we present \textbf{Modality Subspace Activation (MSA)}, a training-free, inference-time framework. 

\subsubsection{Subspace Construction}
To acquire the activation strength of each modality, we first leverage a small set of anchor samples, which are strictly disjoint from the evaluation sets to prevent data leakage, to construct modality-specific feature subspaces.
For each modality $M_i \in \{V, A, T\}$, we compute its activation vector $\mathbf{h}(\{M_i\})$, which is the last hidden state of the model under the single-modality input $M_i$.
By aggregating the activation vectors generated by $N$ samples at the current layer, we construct the raw activation matrix of modality $M_i$, denoted as $H_{M_i}=[\mathbf{h}^1(\{M_i\}), \mathbf{h}^2(\{M_i\}), \dots, \mathbf{h}^N(\{M_i\})]^\top \in \mathbb{R}^{N \times d_{\text{model}}}$, where $d_{\text{model}}$ represents the hidden state size of the model. 
Importantly, to remove static background shifts before applying Singular Value Decomposition (SVD), we perform sample-wise mean-centering on $H_{M_i}$ to derive the centered activation matrix $\hat{H}_{M_i}$:

\begin{equation}
    \begin{aligned}
    \label{eq:mean_centering}
    &\hat{H}_{M_i} = H_{M_i} - \mathbf{1}_N \boldsymbol{\mu}_{M_i}^\top, \quad 
    \\
    &\text{where} \quad \boldsymbol{\mu}_{M_i} = \frac{1}{N} \sum_{j=1}^{N} H_{M_i}[j, :]^\top \in \mathbb{R}^{d_{\text{model}}},
    \end{aligned}
\end{equation}

where $\mathbf{1}_N \in \mathbb{R}^{N \times 1}$ is a column vector of ones, and $\boldsymbol{\mu}_{M_i}$ denotes the feature-wise mean vector across all $N$ samples.

Subsequently, we apply SVD to $\hat{H}_{M_i}$:
\begin{equation}
    \hat{H}_{M_i} = U_{M_i} \Sigma_{M_i} V_{M_i}^T.
\end{equation}
We truncate the top $k$ right singular vectors with the largest singular values, denoted as $\hat{V}_{M_i}=V_{M_i}[:, :k] \in \mathbb{R}^{d_{\text{model}} \times k}$, to serve as the orthogonal basis for the feature activation subspace of modality $M_i$.

\subsubsection{Inference-Time Activation Strength Perception}
During full-modality inference, when the model processes a sample and generates the last hidden state $\mathbf{h}$ prior to the language model head, MSA dynamically perceives the activation strength of each modality. 
The coordinate projection vector of $\mathbf{h}$ onto the modality $M_i$ subspace is formulated as:
\begin{equation}
    \text{proj}_{M_i} = \hat{V}_{M_i} \hat{V}_{M_i}^T \mathbf{h}
\end{equation}
We then employ the $L_2$ norm of the projection vector to quantify the activation strength $a_{M_i}$ of each modality:
\begin{equation}
    a_{M_i} = \Vert \text{proj}_{M_i} \Vert_2 = \Vert \hat{V}_{M_i} \hat{V}_{M_i}^T \mathbf{h} \Vert_2
    \label{action_strength}
\end{equation}
To pinpoint the stronger and the weaker modality, we sort the computed activation strengths in an ascending order:
\begin{equation}
    [a_{\text{weak}}, a_{\text{mid}}, a_{\text{strong}}] = \text{Sort}(\{a_{\text{visual}}, a_{\text{audio}}, a_{\text{text}}\})
\end{equation}
where $a_{\text{weak}} \le a_{\text{mid}} \le a_{\text{strong}}$. 
The projection vectors corresponding to the minimum and maximum elements are explicitly designated as $\text{proj}_{\text{weak}}$ and $\text{proj}_{\text{strong}}$, respectively.

\subsubsection{Dual-Direction Geometric Regulation}
Upon identifying an over-activated(stronger) modality, MSA executes a training-free dual-direction geometric regulation. 
We introduce a balancing factor $\lambda \in [0, 1]$ to govern this adaptive intervention.
The runtime hidden state $\mathbf{h}$ is online rectified to $\mathbf{h}^*$ as follows:
\begin{equation}
    \mathbf{h}^* = \mathbf{h} + \lambda \cdot \text{proj}_{\text{weak}} - (1 - \lambda) \cdot \text{proj}_{\text{strong}}
    \label{msa_enhanced}
\end{equation}
Through the restructuring, the regulation mechanism effectively suppresses the stronger modality that leads the model's decisions, while simultaneously boosting the weaker modality. 
The enhanced hidden state $\mathbf{h}^*$ is subsequently forwarded to the language model head for the token prediction of the final choice. 
By regulating the hidden state at the inference-time, MSA successfully re-establishes the dependency of final decisions on proper perceptual modalities, thereby mitigating the phenomenon of perceptual-decision misalignment.



\section{Experiment}

\begin{table*}[htbp]
\centering
\caption{\textbf{Main performance under various modality perturbations.} Dark red (\best{1st}), medium red (\second{2nd}), and light red (\third{3rd}) highlight top performances in each column. Acc. stands for Accuracy (\%). Acc.\,(MSA) and CMS\,(MSA) denote metrics evaluated with MSA. Avg.\,Perturb. indicates the average over all three modality perturbation settings.}
\label{tab:main_results}
\resizebox{\textwidth}{!}{%
\begin{tabular}{l c c *{3}{c c}l c c}
\toprule
\multirow{2}{*}{\centering Model} & \multicolumn{2}{c}{\textit{Normal}} &
\multicolumn{2}{c}{\textit{No-Visual}} & \multicolumn{2}{c}{\textit{No-Audio}} & \multicolumn{2}{c}{\textit{No-Visual\&Audio}} &
\multicolumn{3}{c}{Avg. Perturb.}\\
\cmidrule(lr){2-3} \cmidrule(lr){4-5} \cmidrule(lr){6-7} \cmidrule(lr){8-9} \cmidrule(lr){10-12}
 & Acc. & Acc.(MSA) & Acc. & CMS & Acc. & CMS & Acc. & CMS &
Acc. & CMS & CMS(MSA) \\
\cmidrule(lr){1-1} \cmidrule(lr){2-3} \cmidrule(lr){4-9} \cmidrule(lr){10-12}
Qwen2.5-Omni-3B
& 30.94 & 25.57 & 
33.39 & 30.47 & 
28.66 & 12.11 & 
20.85 & 35.69 & 
27.63($\downarrow$3.31) & 26.09 & \second{52.28}  \\
Qwen2.5-Omni-7B
& \third{49.35} & \second{55.70} & 
44.79 & 33.83 & 
\third{42.53} & 15.75 & 
29.97 & 45.79 & 
39.09($\downarrow$10.26) & 31.79 & \best{52.36}  \\
Qwen3-Omni-30B-A3B-Instruct
& \best{57.33} & \best{65.15} & 
\best{52.61} & 22.77 & 
\best{53.26} & \third{21.03} & 
\best{43.97} & 35.41 & 
\best{49.95}($\downarrow$7.38) & 26.40 &  30.58  \\
OmniVinci
& 35.34 & 22.80 & 
35.34 & 26.79 & 
31.76 & 13.35 & 
28.01 & 29.20 & 
31.70($\downarrow$3.64) & 23.12 & 34.55    \\
Interactive-Omni-4B
& 47.88 & 40.07 & 
40.55 & \best{43.25} & 
\second{42.35} & 17.13 & 
27.36 & \second{57.49} & 
36.75($\downarrow$11.13) & \second{39.29} & 45.58   \\
Interactive-Omni-8B
& 46.42 & \third{45.60} & 
\second{47.39} & \second{40.22} & 
42.51 & \second{21.72} & 
\second{34.85} & \third{54.58} & 
\third{41.59}($\downarrow$4.83) & \third{38.84} & 47.65    \\
MiniCPM-4.5
& \second{51.95} & 40.23 & 
\third{45.44} & \third{38.90} & 
\second{47.07} & \best{26.72} & 
\third{34.53} & \best{59.62} & 
42.35($\downarrow$9.60) & \best{41.74} & \third{49.72}    \\

\bottomrule
\end{tabular}%
}
\end{table*}

\subsection{Experimental Settings}
\subsubsection{CausalMSBench}
To create an evaluation dataset that minimizes language priors while strictly enforcing multi-modal dependency, we built the \textbf{CausalMSBench} (617 samples) based on JointAVBench~\cite{chao2026jointavbench}, DailyOmni~\cite{zhou2026dailyomni}, and samples we collected and annotated ourselves. 
We constructed this dataset both to fulfill our core evaluation requirements and to introduce greater diversity while avoiding trivial questions.
The construction method is described specifically in the supplementary materials.

\subsubsection{Subspace Construction Samples Source}
The set of anchor samples applied to acquire the activation strength of each modality should be strictly disjoint from the evaluation sets to prevent data leakage. 
Therefore, we adopt the SocialOmni~\cite{xie2026socialomni} dataset, which focuses on three dynamic social interaction dimensions in conversations, namely who is speaking, when interruptions occur, and how they occur, making it highly suited for our multi-modal content understanding scenario. 
We randomly select 500 samples from this dataset to construct the activation matrices.

\subsubsection{Other Settings}
For our experimental models, we evaluate popular Omni-LLMs, including Qwen2.5-Omni~\cite{xu2025qwen25omnitechnicalreport}, Qwen3-Omni~\cite{xu2025qwen3omnitechnicalreport}, InteractiveOmni~\cite{tong2025interactiveomniunifiedomnimodalmodel}, OmniVinci~\cite{ye2025omnivincienhancingarchitecturedata}, and MiniCPM-4.5~\cite{yu2025minicpmv45cookingefficient}.
The samples applied to construct the activation matrices are 500 random samples from SocialOmni.
In the MSA method, we select the top $k$ principal components of the subspace. 
By default, we set $k=64$, which captures the vast majority of the cumulative singular value energy.
In our setup, where the hidden state dimension $d_{\text{model}}$ ranges from 2,000 to 4,000, the condition $k \ll N < d_{\text{model}}$ holds, ensuring that $N=500$ anchor samples provide sufficient statistical support for stable basis estimation.
To evaluate model behaviors under cross-modal degradation, we apply three primary perturbation protocols: \textit{No-Visual}, \textit{No-Audio}, and \textit{No-Visual\&Audio}, corresponding to the omission of visual input, audio input, and both modalities, respectively.
We used Accuracy, CMS as evaluation metrics.

\vspace{-12pt}
\subsection{Overall Results}
\vspace{-12pt}

As shown in the results in Table~\ref{tab:main_results}, we conducted experiments on multiple Omni-LLMs to compare the performance of each model in terms of Accuracy and CMS metrics before and after applying MSA.
For different modality perturbations, we can see that virtually all models follow the CMS pattern: No-Visual\&Audio > No-Visual > No-Audio. 
This indicates that even minimizing language priors and strictly enforcing multi-modal dependency, the absence of the audio modality has little impact on models' outputs in the evaluated models.
Across different models, based on the CMS metric, we observe that MiniCPM-4.5 exhibits the highest sensitivity to modalty perturbations, while OmniVinci exhibits the lowest. 

Based on accuracy under normal conditions, Qwen3-Omni-30B-A3B-Instruct appears to perform the best, with a response accuracy rate of 57.33\%, which is higher than that of MiniCPM-4.5.
However, our CMS metric indicates that, after modality perturbation, the output variation of Qwen3-Omni-30B-A3B-Instruct is not as significant as that of MiniCPM-4.5.
Furthermore, compared to MiniCPM-4.5, Interactive-Omni-4B displays a greater drop in accuracy after modality perturbation, amounting to 11.13\%.
However, according to the CMS metric, the degree of misalignment between perception and decision-making in Interactive-Omni-4B is actually lower than that of MiniCPM-4.5.
Therefore, in evaluating these models, compared to the traditional accuracy metric, our CMS offers another diagnostic perspective.

\begin{figure}[H]
    \centering
    \includegraphics[width=0.75\columnwidth]{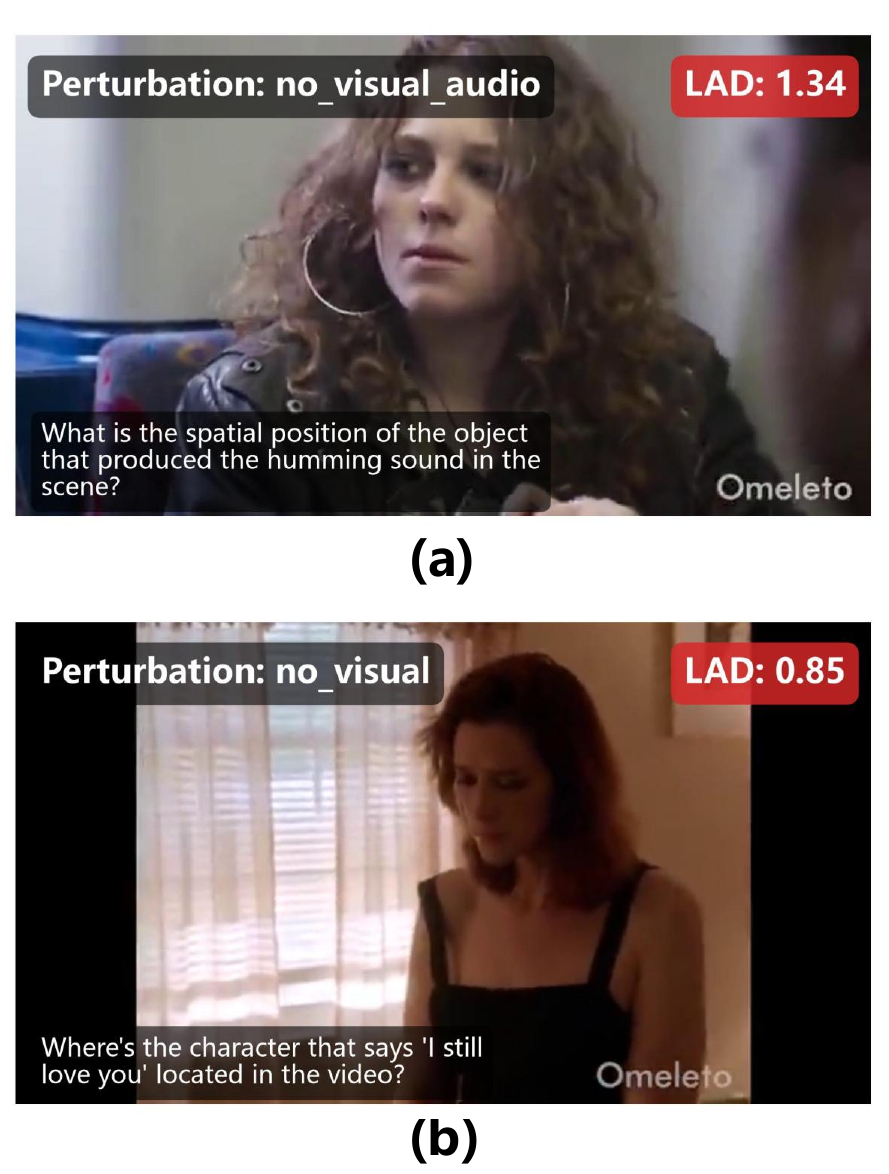}
    \caption{\textbf{Representative Cases of Modality-Sensitivity Diagnostics in Interactive-Omni-4B}. In both cases (a) and (b), the predicted textual answers remain correct and unchanged before and after modality perturbations. However, the high LAD values ($1.34$ and $0.85$) effectively capture significant underlying shifts in the internal multi-choice space.}
    \label{fig:case_study}
\end{figure}

By comparing the Accuracy and CMS metrics before and after applying MSA, we can see that the CMS of all models has improved. 
This indicates that MSA does indeed boost the impact of modality perturbation on model outputs.
By amplifying the impact of modality perturbations on model outputs, MSA makes it more difficult for models to rely solely on the stronger modality when reasoning over full-modality inputs on CausalMSBench.
Consequently, applying MSA yields distinct accuracy changes across models in the Normal (full-modality input) setting. 
Most models experience a marginal accuracy decline, whereas Qwen3-Omni-30B-A3B-Instruct and Qwen2.5-Omni-7B achieve slight improvements. 
We reason that MSA enhances multi-modal fusion in certain models, while for others, it restricts their potential reliance on shortcut learning.

\subsection{Case Study}
As shown in Figure~\ref{fig:case_study}, for both samples (a) and (b), Interactive-Omni-4B yields correct predictions under full-modality conditions and maintains an ARR of 1.0 under No-Visual\&Audio and No-Visual perturbations.
However, the LAD values for (a) and (b) are $1.34$ and $0.85$, respectively, indicating that even when the discrete predictions remain unchanged after modality perturbations, the internal continuous logit distributions over candidate options undergo drastic shifts.
This further demonstrates the necessity of our proposed LAD as a complementary diagnostic metric.

\begin{figure}[h]
    \centering
    \includegraphics[width=0.8\columnwidth]{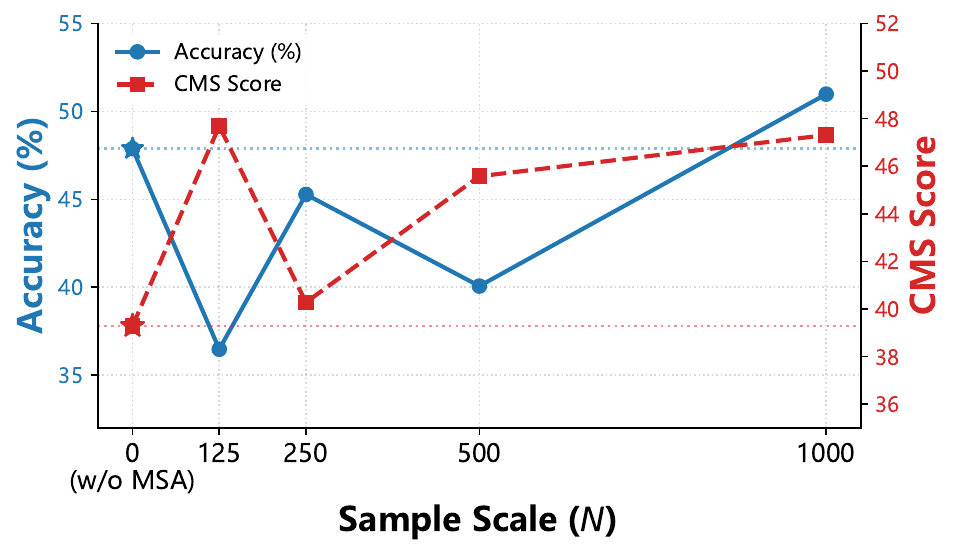}
    \caption{\textbf{Effect of the activation sample scale $N$ on model performance.} The horizontal axis denotes the sample scale $N$ ($N=0$ represents the unsteered baseline without MSA), while the left and right vertical axes indicate Accuracy (\%) and CMS (\%), respectively.}
    \label{fig:sample_scale_efficiency}
\end{figure}

\vspace{-12pt}
\subsection{The Analysis and Discussion of MSA}

\subsubsection{Sample Scale Sensitivity in Activation Matrix Construction}
To evaluate how the sample size $N$ used for constructing modality activation matrices affects experimental results, we vary $N \in \{125, 250, 500, 1000\}$ and compare these variants against the baseline ($N=0$, i.e., without MSA).
As depicted in Figure~\ref{fig:sample_scale_efficiency}, small sample sizes (e.g., $N=125$) yield suboptimal performance, which we attribute to the noisy singular directions extracted from insufficient activation samples. 
As $N$ scales up to $1000$, the extracted SVD subspaces become stable and representative. 
However, larger sample sizes incur higher memory footprints and computational overhead during matrix construction. 
Therefore, we adopt $N=500$ as an optimal choice, striking a balance between sample efficiency and representation quality.

\subsubsection{Sensitivity Analysis on Intervention Strength}
%
\begin{figure}[h]
    \centering
    \includegraphics[width=0.8\columnwidth]{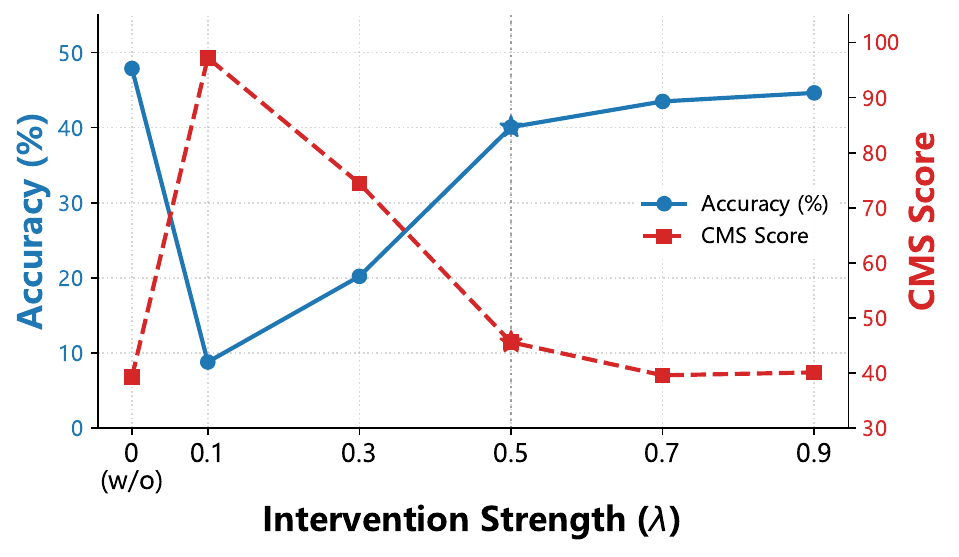}
    \caption{\textbf{Sensitivity Analysis on intervention strength $\lambda$.} The horizontal axis denotes $\lambda$, while the left and right vertical axes indicate Accuracy (\%) and CMS (\%), respectively.}
    \vspace{-6pt}
    \label{fig:lambda_sensitivity}
\end{figure}
To validate the choice of the balancing factor $\lambda$, we examine Accuracy and the CMS scores across $\lambda \in [0, 0.9]$, as shown in Figure~\ref{fig:lambda_sensitivity}. 
When $\lambda$ is small (e.g., $\lambda=0.1$), excessive misalignment causes task Accuracy to plummet to $8.79\%$ despite an inflated CMS score. 
As $\lambda$ increases, the model rapidly recovers its reasoning stability. 
Setting $\lambda=0.5$ provides an optimal Pareto trade-off, preserving reliable accuracy ($40.07\%$) while maintaining effective representation enhancement.

\subsubsection{Different Enhancement Strategies}
\begin{figure}[h]
    \centering
    \includegraphics[width=0.8\columnwidth]{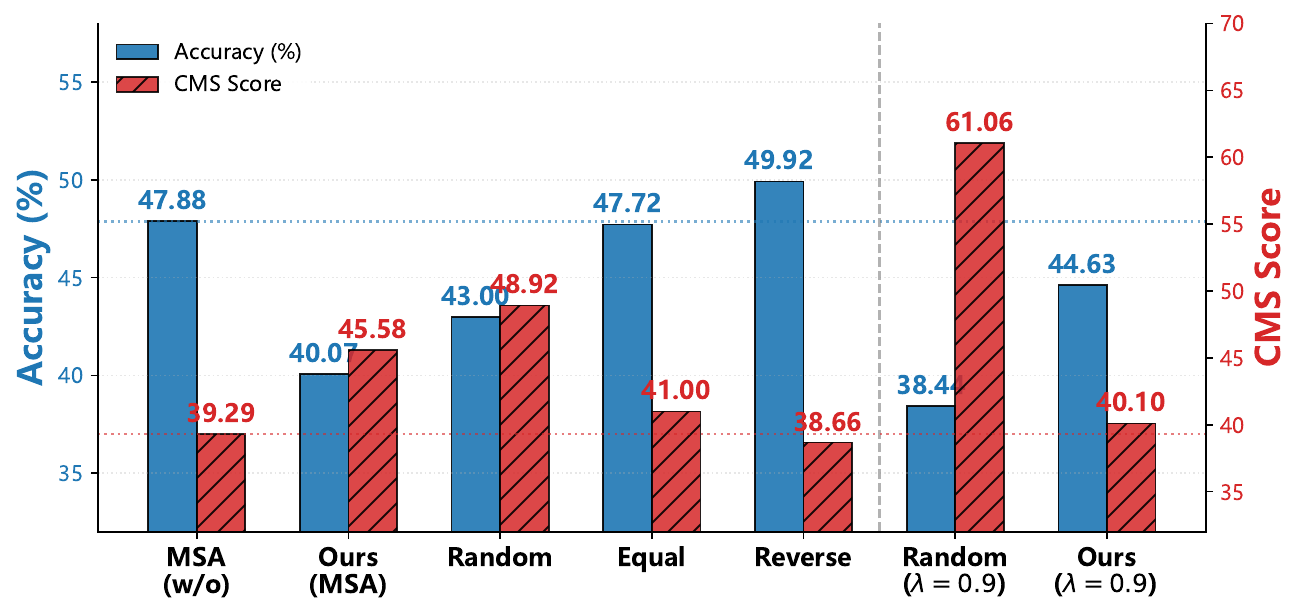}
    \caption{\textbf{Efficacy Comparison of Enhancement Strategies and High Intervention Strengths.} Blue bars (left axis) and red hatched bars (right axis) denote Accuracy (\%) and CMS (\%), respectively. The vertical dashed line separates default settings from the extreme balancing factor ($\lambda=0.9$).}
    \vspace{-6pt}
    \label{fig:enhance_strategy}
\end{figure}
As shown in Figure~\ref{fig:enhance_strategy}, we evaluate the efficacy of different steering strategies, with additional variants detailed in the Supplementary Material. 
Under the default setting ($\lambda=0.5$), \texttt{Random} achieves performance comparable to \texttt{Ours (MSA)}, as a lower intervention strength conceals the drawbacks of unguided enhancement.
To verify this, escalating the intervention strength to $\lambda=0.9$ reveals that \texttt{Random} yields an inflated CMS spike ($61.06$) at the cost of a severe Accuracy drop ($38.44\%$). In contrast, \texttt{Ours (MSA)} maintains robust Accuracy ($44.63\%$), confirming that lower intervention strengths mask the inherent weakness of unguided enhancement.
Regarding other variants, \texttt{Equal} enhances all modality uniformly, whereas \texttt{Reverse} suppresses weak modality while amplifying dominant ones, thereby further exacerbating modality imbalance. 
Consequently, both strategies yield lower CMS scores than \texttt{Ours (MSA)}.
Notably, while they exhibit slightly higher accuracy, this gain may come at the cost of overrelying on shortcuts in dominant modalities.

\section{Conclusion}
In this work, we investigate Perceptual-Decision Misalignment (PDM) in Omni-LLMs, where models yield invariant predictions even when indispensable modalities are omitted. We curate CausalMSBench under strict multi-modal dependency and evaluate model behaviors using Answer Retention Rate (ARR) and Logit Angular Discrepancy (LAD). To restore modal reliance, we present Modality Subspace Activation (MSA), a training-free mechanism that geometrically suppresses dominant modality activation while boosting weaker ones on hidden representations. Quantitative evaluations confirm that MSA enhances causal modality sensitivity and reduces reliance on shortcut learning.


\newpage
\bibliography{aaai2027}


\end{document}